\documentclass[letterpaper, 10 pt, conference]{ieeeconf} 
\IEEEoverridecommandlockouts
\usepackage{amsmath}
\usepackage{amssymb}

\newcommand{\contime}   {t}                      

\newcommand{\x}       {x}                      
\newcommand{\bx}      {\mathbf{\x}}            
\newcommand{\bxdot}   {\dot{\mathbf{\x}}}         
\newcommand{\bX}      {\mathbf{\MakeUppercase{\x}}} 
\newcommand{\dimX}    {\mathcal{M}}            

\newcommand{\uv}       {u}                      
\newcommand{\bu}      {\mathbf{\uv}}            

\newcommand{\y}       {y}                   
\newcommand{\by}      {\mathbf{\y}}            
\newcommand{\dimY}    {\mathcal{P}}            

\newcommand{\nd}      {\mathit{n}}             
\newcommand{\ndt}      {\mathit{a}}             
\newcommand{\ndv}      {\mathit{b}}             
\newcommand{\nde}      {\mathit{c}}             
\newcommand{\Nd}      {\mathcal{\MakeUppercase{\nd}}}  
\newcommand{\Ndtrain}      {\mathfrak{\MakeUppercase{\ndt}}}  
\newcommand{\Ndvalid}      {\mathfrak{\MakeUppercase{\ndv}}}  
\newcommand{\Ndeval}      {\mathfrak{\MakeUppercase{\nde}}}  
\newcommand{\nt}      {\mathit{t}}             
\newcommand{\Nt}      {\mathcal{\MakeUppercase{\nt}}}  
\newcommand{\Nts}      {\mathcal{{T_s}}}  

\newcommand{\kn}      {\mathit{k}}       

\newcommand{\Xspace}    {\mathcal{X}}           
\newcommand{\Uspace}    {\mathcal{U}}           
\newcommand{\Rspace}    {\mathbb{R}}           
\newcommand{\Cspace}    {\mathbb{C}}           
\newcommand{\Hspace}    {\mathcal{H}}          

\newcommand{\Fn}        {F}                  
\newcommand{\bFn}       {\mathbf{\Fn}}       

\newcommand{\Koop}    {\mathcal{K}}     
\newcommand{\Koopd}    {\mathbf{K}}    
\newcommand{\KEigF}   {\phi}          
\newcommand{\bKEigF}   {\boldsymbol{\KEigF}}          
\newcommand{\KEigV}   {\lambda}        
\newcommand{\KEigVReal}   {\mu}        
\newcommand{\KEigVImag}   {\omega}        

\newcommand{\gn}        {g}                  

\newcommand{\q}    {q}     
\newcommand{\qdot}    {\dot{q}}     
\newcommand{\qddot}    {\ddot{q}}     

\newcommand{\vf}    {\upsilon}     
\usepackage[utf8]{inputenc}
\usepackage{cite}
\usepackage{amsmath,amssymb,amsfonts}
\usepackage{textcomp}
\usepackage{xcolor}
\usepackage{amsmath}
\usepackage{tabularx}
\usepackage{amsfonts}
\usepackage{amssymb}
\usepackage{graphicx}\graphicspath{{figures/}} 
\usepackage{caption}
\usepackage{subcaption}
\usepackage[acronym]{glossaries}
\usepackage{siunitx}
\usepackage{algorithm,algorithmic}
\usepackage{amssymb}
\usepackage{pifont}
\makeatletter\let\@noitemerr\relax\makeatother
\usepackage{setspace}
\usepackage{float}
\usepackage{caption}
\usepackage{verbatim}
\let\labelindent\relax 
\usepackage[inline,shortlabels]{enumitem}\setlist[enumerate]*{label=(\roman*)}
\usepackage{multirow} 
\usepackage{rotating}
\usepackage{color}
\usepackage{grffile} 
\usepackage{rotating}
\usepackage{array}
\newcolumntype{H}{>{\setbox0=\hbox\bgroup}c<{\egroup}@{}}
\usepackage{makecell}
\usepackage{url}

\newcommand{\eg}{\textit{e.g.,}~} %
\newcommand{\ie}{\textit{i.e.,}~} %
\def\figurename{Fig.}
\def\equationname{Eq.} 

\newcommand*{\sref}[1]{\S\ref{s:#1}}            
\newcommand*{\tref}[1]{\tablename~\ref{t:#1}}   
\newcommand*{\fref}[1]{\figurename~\ref{f:#1}}  
\newcommand*{\eref}[1]{\equationname~(\ref{e:#1})}            

\makeatletter\newcommand{\manuallabel}[2]{\def\@currentlabel{#2}\label{#1}}\makeatother

\usepackage[percent]{overpic} 
\usepackage{xargs}                      

\usepackage{amsmath,array,graphicx}
\usepackage{kantlipsum}

\def\BibTeX{{\rm B\kern-.05em{\sc i\kern-.025em b}\kern-.08em
    T\kern-.1667em\lower.7ex\hbox{E}\kern-.125emX}}

\title{\LARGE \bf
Deep Koopman with Control: Spectral Analysis of Soft Robot Dynamics
}

\author{Naoto Komeno$^{1,*}$, Brendan Michael$^{1,*}$,  Katharina K{\"u}chler$^{1,2}$, Edgar Anarossi$^{1}$ and Takamitsu Matsubara$^{1}$%
\thanks{$^{1}$Robot Learning Lab, Nara Institue of Science and Technology, Japan}%
\thanks{$^{2}$Institute of Knowledge Based Systems Group, RWTH Aachen University, Germany}%
\thanks{$^{*}$These authors contributed equally to this work}%
\thanks{This work was supported by JSPS KAKENHI Grant Numbers JP19H01124 and JP22J11687.}
}

\begin{document}
\maketitle
\thispagestyle{empty}
\pagestyle{empty}

\begin{abstract}
Soft robots are challenging to model and control as inherent non-linearities (\eg elasticity and deformation), often requires complex explicit physics-based analytical modelling (\eg a priori geometric definitions). While machine learning can be used to learn non-linear control models in a data-driven approach, these models often lack an intuitive internal physical interpretation and representation, limiting dynamical analysis. 
To address this, this paper presents an approach using Koopman operator theory and deep neural networks to provide a global linear description of 
the non-linear control systems.
Specifically, by globally linearising dynamics, the Koopman operator is analyzed using spectral decomposition to  characterises important physics-based interpretations, such as functional growths and oscillations. Experiments in this paper demonstrate this approach for controlling non-linear soft robotics, and shows model outputs are interpretable in the context of spectral analysis. 
\end{abstract}

\section{Introduction}

Linear control theory is well suited to developing interpretable control frameworks, through exploration of the spectral components, \ie eigenvectors and eigenvalues, of the associated dynamical system. Spectral analysis can help determine system stability \cite{michiels2014stability}, or provide additional insight for techniques such as filtering \cite{ham1983observability}.  However, application to \textit{non-linear} systems is ill-suited, due to the absence of a linear evolution of the dynamics, resulting in sub-optimal solutions and poor control applications.

In particular, soft robots with non-linear properties (\eg elasticity) suffer from difficulties in both modelling and control (\eg unpredictable behaviour due to the high degree of freedom). As such, predictive control often requires physics based modelling \cite{grazioso2019geometrically}, including analytical descriptions of the geometries  \cite{sadati2021tmtdyn}. However, dynamical analysis of non-linear systems remains challenging, and is generally limited to systems with closed-form derivations (\eg double pendulums \cite{yu1998analysis}).

As an alternative to analytical modelling, machine learning can be employed to learn predictive models of the non-linear dynamical system, solely through observations of the environment. While there exists a large body of work exploring machine learning methods for soft robots \cite{kim2021review}, a common limitation is the lack of interpretability and explainability of models \cite{heuillet2021explainability}. Specifically, in the context of learning predictive models for control, black-box machine learning systems \cite{rudin2019stop} may only locally linearise the dynamics \cite{huang2003neural}, thereby not capturing intrinsic important global physical properties of the system. This both limits dynamics analysis of the learnt model, and reduces confidence in model generalisability.

\begin{figure}[t]
\begin{center}
\includegraphics[width=\hsize]{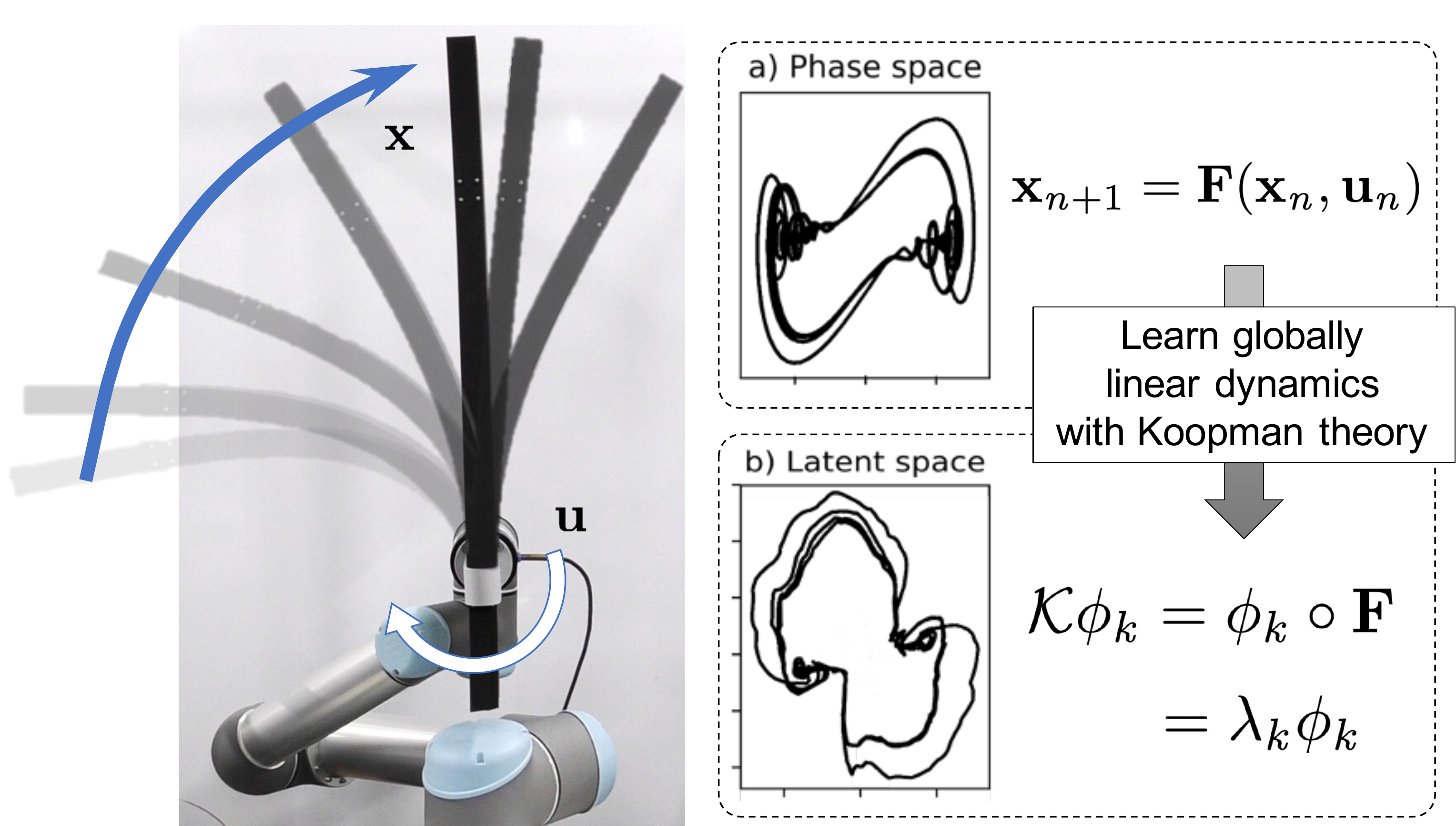}
\caption{\label{f:senzai} Stabilization of a flexible polyurethane arm. (a) Complex non-linear robot dynamics are mapped to set of (b) linearised latent dynamics (polar co-ordinates), via Koopman operator theory, becoming amenable to spectral analysis.}
\vspace{-5mm}
\end{center}
\end{figure}

To address this, this paper proposes an approach to controlling and interpreting non-linear soft robotics by learning \textit{globally linearized dynamics models} via \textit{Koopman operator theory} \cite{koopman1931hamiltonian}, and applying this to model predictive controllers.  
Specifically, \textit{Deep Koopman Networks} (DKN) \cite{lusch2018deep} is used to learn parsimonious dynamics models with control, that expresses linear dynamics interpretations in the context of spectral analysis. 
Prior work applying Koopman theory to robotics \cite{haggerty2020modeling,abraham2019active,han2020deep,shi2022deep} is limited to improving prediction and control performance, without considering the interpretability of the model. 
This paper presents the first evidence of DKN in soft robotics for non-linear control with dynamics analysis contextualised within the linear control domain.




Experiments apply the approach to a soft robot system (soft inverted pendulum) for both modeling the dynamics, and stabilisation control. Results show an improved control performance over standard deep learning for a soft flexible stabilisation task, and models display clear physical interpretations of the dynamic system.
\section{Background}

Koopman operator theory \cite{koopman1931hamiltonian} is a dynamical systems formulation, that provides a global description of non-linear dynamics, in terms of the \textit{linear evolution} of a set of \textit{observable functions}. Specifically, instead of attempting to model the non-linear dynamical state space (\eg positions and velocities), Koopman operator theory alternatively describes the dynamical system in terms of the linear evolution of a set of state-dependent observable functions, forming a latent linear dynamical system \cite{mauroy2020koopman}. 












\subsection{Koopman operator theory}

To present the role of Koopman operator theory in dynamical systems analysis, a discrete-time dynamical system is first formulated:
\begin{equation}
    \bx_{\nd+1} = \bFn(\bx_{\nd}),
    \label{e:disc_dyn}
\end{equation}
where $\nd$ is the time-index,  $\bFn: \Xspace \mapsto \Xspace$ is the flow map, and $\Xspace$ is a finite-dimensional metric state-space \cite{mauroy2020koopman}, often assumed to be a smooth manifold \cite{kutz2016dynamic} (\eg Euclidean, $\Xspace \subseteq \Rspace^{\dimX}$).  

Analysis and modelling of this non-linear $\bFn$ can be challenging. Instead, a reasonable approach is to find coordinate transformations to map from the non-linear dynamics to a latent linear dynamical system. Koopman operator theory does this by describing the linear evolution of measurement functions of the non-linear state \cite{kutz2016dynamic}: 
\begin{equation}
    {\Koop} \gn ({\bx}_{\nd}) = \gn({\bFn}({\bx}_{\nd})) = \gn(\bx_{\nd+1}),
    \label{e:Koopman_flow}
\end{equation}
where ${\Koop}_{\contime}$ is the \textit{Koopman operator}, an infinite-dimensional linear operator, acting on a measurement function $\gn$ of the system (also known as an \textit{observable.}). This observable is a function of the state, \ie $\gn: \Xspace \mapsto \Cspace$, and an infinite set of observable functions is defined on a Hilbert space $\Hspace$ \cite{kutz2016dynamic}. This formulation is often referred to as \textit{lifting} the state variables from the finite non-linear space, to an infinite-dimensional linear one. Given all observables, $\Koop$ controls their evolution, \ie ${\Koop}: \Hspace \mapsto \Hspace$. The Koopman operator is associated with the non-linear state transformation $\bFn$ through the composition ${\Koop} \gn = \gn \circ \bFn, \forall \gn \in \Hspace$ \cite{mauroy2020koopman}. 


\subsection{Koopman theory with control}

Koopman operator theory is also applicable to systems with control inputs, by describing the dynamics of an extended state space of the product of $\Xspace$ and the space of all control sequences $\ell(\Uspace)$ \cite{Korda2018}. By defining the operator on the extended space, the Koopman operator remains autonomous and is equivalent to the operator associated with unforced dynamics \cite{brunton2021modern}. Specifically, considering a non-linear dynamical system with control $\bu \in \Uspace$:
\begin{equation}
    \bx_{\nd+1} = \bFn(\bx_{\nd},\bu_{\nd}),
    \label{e:disc_dyn_control}
\end{equation}
a common generalization is to define the associated Koopman operator as evolving an \textit{uncontrolled} dynamic system defined by the product $\bFn: \Xspace \otimes \Uspace \mapsto \Xspace$ \cite{Korda2018}. 

As the evolution of $\Xspace$ only depends on $\bu_{\nd}$, the control is considered as an additional state variable \cite{mauroy2020koopman}, instead of requiring all control sequences $\ell(\Uspace)$. As such, similar to \eref{Koopman_flow}, the Koopman operator is described as \cite{mauroy2020koopman}:
\begin{equation}
    {\Koop} \gn ({\bx}_{\nd},{\bu}_{\nd}) = \gn(\bFn(\bx_{\nd},\bu_{\nd}),\bu_{\nd+1}). 
    \label{e:Koopman_flow_control}
\end{equation}

In contrast to 
linear predictors \cite{Korda2018},
$\bu$ is also lifted through observables $\gn$ alongside the state $\bx$.

Applications of Koopman operator theory to domains with control are are fast appearing in the literature, and generally focus on strategies for choosing suitable observable functions \cite{Brunton2016} or optimal control \cite{kaiser2017data,Korda2018}. For a general review of Koopman based control frameworks, please see \cite{bevanda2021koopman}. With specific regard to robotics control applications, prior work has investigated finding observables \cite{Shi2021}, LQR \cite{mamakoukas2019local,haggerty2020modeling,shi2022deep} or MPC \cite{folkestad2020episodic,bruder2020data,sotiropoulos2021dynamic} control, active learning \cite{abraham2019active}, and shared human-robot control \cite{broad2020data}.

\subsection{Spectral analysis and finite approximations of the Koopman operator}

Analysis of this infinite-dimensional linear operator is challenging. However, finite spectral properties of the operator are of great importance, as they can outline global properties of the dynamics \cite{mezic2005spectral}. Specifically, $\Koop$ can be spectrally decomposed into Koopman eigenfunctions ${\KEigF}_{\kn}(.) \in \Hspace \backslash \{$$0$$\}$ \cite{mauroy2020koopman}, with corresponding Koopman eigenvalue $\KEigV_{\kn} \in \Cspace$), which satisfies: 
\begin{equation}
    \Koop \KEigF_{\kn}  = \KEigF_{\kn}  \circ \bFn = \KEigV_{\kn} \KEigF_{\kn}. 
    \label{e:koopman_eq}
\end{equation}

This Koopman eigenfunction is a linear intrinsic measurement coordinate on which measurements are evolved with a linear dynamical system \cite{kutz2016dynamic}. As such, spectral analysis of this can  provide physical intuition of the dynamical system under investigation.

While there exists a large body of work on finding analytical representations of eigenfunctions with knowledge of the dynamical system  \cite{mauroy2020koopman,Brunton2016}, data-driven approximations are increasing prevalent. This is due to the complexity and uncertainty in finding eigenfunctions, and the modern increase in computational power for modelling, and availability of large datasets.

Generally, approximations are often made using the \textit{dynamic mode decomposition} algorithm \cite{rowley2009spectral,kutz2016dynamic}, whereby spectral components of a linear transition matrix are estimated via  matrices of state measurements. Specific implementations that incorporate Koopman operator theory to handle non-linear transitions include dictionary \cite{li2017extended} or deep learning \cite{takeishi2017learning} for finding observables  or eigenfunctions \cite{lusch2018deep}, or utilizing approaches such as time-delay embeddings \cite{kamb2020time}.

\section{Approach}

\subsection{Deep Koopman Network}
\label{s:DKN}

A promising approach to learning the linearisation via eigenfunctions and the latent linear dynamics \eref{koopman_eq} 
for autonomous uncontrolled systems,
is the \textit{Deep Koopman Network} (DKN) approach \cite{lusch2018deep}. 
In this, a deep autoencoder framework is used to find intrinsic latent coordinates 
$\by=\bKEigF(\bx)$
approximating the Koopman eigenfunctions $\bKEigF: \Rspace^{\dimX} \mapsto \Rspace^{\dimY}$, and associated linear dynamical system $\by_{\nd+1}=\Koopd \by_{\nd}$.
This is achieved by using an autoencoder to learn an encoder $\bKEigF$ and decoder $\bKEigF^{-1}$, and inner layers to learn the  linear dynamics $\Koopd$. To learn parsimonious models, continuous spectra dynamics are captured by parameterising $\Koopd$ by an auxiliary layer, predicting eigenvalues as a function of $\by$.

An intuitive design constraint for DKN,  is to learn latent coordinates $\by$ which have complex radial symmetry \cite{lusch2018deep}, as exponentials can be seen as eigenfunctions of a differential operator \cite{peter2013generalized}. As such, for each eigenfunction $\bKEigF_{\kn}$, the associated eigenvalue is given as a complex pair, representing circular motion in the latent space. Specifically, $\KEigV_{\kn} = \KEigVReal_{\kn} \pm i \KEigVImag_{\kn}$, where $\KEigVReal_{\kn}$ denotes the growth/decay, and $\KEigVImag_{\kn}$ the oscillation, and for each complex eigenvalue pair an associated linear operator $\Koopd$ 
with sampling time $\Delta \contime$
is formed as:
\begin{equation}
{\Koopd}(\KEigVReal_{\kn},\KEigVImag_{\kn}) = 
    \exp(\KEigVReal_{\kn} \Delta \contime)    \begin{bmatrix}
   \cos(\KEigVImag_{\kn} \Delta \contime ) & -\sin(\KEigVImag_{\kn} \Delta \contime ) \\
   \sin(\KEigVImag_{\kn} \Delta \contime )  & \cos(\KEigVImag_{\kn} \Delta \contime )
\end{bmatrix}.
\end{equation}
Given this design constraint, $\bKEigF$, $\bKEigF^{-1}$, and $\Koopd$ are learnt using measurement data.

Here, we regard control input $\bu$ as one of the state variables for non-linear control systems so that DKN learns non-linear control dynamics.
Previous studies using deep-learning-based learning of Koopman operators \cite{han2020deep,shi2022deep} sacrifice interpretability by assuming bi-linear systems for \eref{disc_dyn_control} to achieve optimal control, but our approach realizes both interpretability and controllability.
For the latent coordinates $\by=\bKEigF([\bx,\bu])$, datasets are given in the DMD format of delayed snapshot matrices \cite{kutz2016dynamic}
${\bX} = [[{\bx}_0,{\bu}_0],\dots,[{\bx}_{{\Nd}-1},{\bu}_{{\Nd}-1}]]$ and outputs ${\bX}' = [[{\bx}_1,{\bu}_1],\dots,[{\bx}_{{\Nd}},{\bu}_{{\Nd}}]]$. Eigenfunctions and linear dynamics are learnt by minimising a set of loss functions:
\begin{enumerate*}
    \item Reconstruction loss: $||{\bx}_{\nd} - \bKEigF^{-1}(\bKEigF([{\bx}_{\nd},{\bu}_{\nd}]))||$,
    \item Linear dynamics loss: $||\bKEigF([{\bx}_{\nd+1},{\bu}_{\nd+1}]) - {\Koopd}(\bKEigF([{\bx}_{\nd},{\bu}_{\nd}]))||$,
    \item Next-step prediction loss: $||\bx_{\nd+1} -  \bKEigF^{-1}({\Koopd}(\bKEigF([{\bx}_{\nd},{\bu}_{\nd}])))||$.
\end{enumerate*}
In this paper, all components in ${\bx}_n$ are assumed to be measurable.
For specific details regarding implementation and learning  (including initialisations, regularisations and hyperparameters), see \cite{lusch2018deep}.

\subsection{Sampling-based model predictive control}

Given this predictive model of the dynamics, a reasonable approach to controlling a system to a desired state is via a model-based controller. In this study, even in linear latent space, control inputs are given as non-linear system state variables \eref{Koopman_flow_control}. As such, standard linear optimal control (\eg LQR) applied in previous studies  \cite{mamakoukas2019local,haggerty2020modeling,han2020deep,shi2022deep} is inapplicable. Instead, sampling-based nonlinear model predictive control (MPC) is employed to overcome the control non-linearities. This is applied as a model-based control method in nonlinear observation space.

Specifically, MPC uses the  approximation of \eref{disc_dyn_control} from learnt DKN components \sref{DKN}, in combination with a given control input at each time step, to predict the future state. To determine the optimal control to reach the desired target state, a minimization problem is formulated with a cost $c$ computed as the difference between the predicted state and the target state ${\bx}_{target}$. Given this cost, sampling-based optimization using the cross-entropy method (CEM)\cite{cem} is performed.
The optimization problem for MPC is formulated:
\begin{equation}
    u_t = \mathop{\rm arg~min}\limits_{u_t} \sum_{i=t+1,...,t_p} c_i({\bx}_i,{\bx}_{target}),
    \label{e:MPC}
\end{equation}
where $t_p$ is prediction horizon to determine the action that will bring the prediction result closest to the target state. 
\section{Simulation}
\label{s:sim_experiments}
To evaluate the proposed approach for learning non-linear dynamics models that explain intuitive physical aspects, experiments are presented to explore the application of DKN with control. This evaluation is comprised of three constituent parts:
\begin{enumerate*}
    \item learning dynamics of a non-linear system under control, 
    \item analysing the learnt Koopman outputs to elucidate an intuitive understanding of the system, and
    \item applying predictive control.
\end{enumerate*}

\subsection{Simulation experiments}

Initially, an experiment is presented to learn and analyse the dynamics of a non-linear rigid pendulum, similar to that proposed in \cite{lusch2018deep}. While a relatively mundane system, the non-linear pendulum is challenging to model within a parsimonious framework, due to it exhibiting a continuous eigenvalue spectrum (\eg differing oscillation frequency dependant on dynamical state) which often requires a harmonic expansion estimation. 

The dynamics of the pendulum under control \cite{doya2000reinforcement} is entirely described by the three dimensional state ${\bx} = [\q,\qdot,\uv]^\intercal$, where $\q$ is the joint angle, $\qdot$ the joint velocity, and $\uv$ the applied torque. The equations of motion for the rigid pendulum is given as:
\begin{equation}
    \bxdot = [\qdot,\qddot]^\intercal,
    \label{e:rigid_eq}
\end{equation}
where angular acceleration $\qddot$ is:
\begin{equation}
    \qddot = \frac{g}{l} \sin(\q) + \frac{(\uv - \vf \qdot)}{{(m l)}^2},
    \label{e:rigid_dynamics}
\end{equation}
with gravity $g$, rod length $l$, mass $m$ and viscous friction $\vf$.  The aim of this experiment is to learn a predictive model of \eref{rigid_eq} and \eref{rigid_dynamics}, through observations of $\q$, $\qdot$, and $\uv$. 

\subsection{Generating data}
\label{s:generating_data}
 Trajectories of motion are generated as samples from which to learn dynamics models. For this, \eref{rigid_dynamics} is solved for $\Nts$ time-steps with a forth-order Runge-Kutta method. From these $\Nts$ timesteps, $\Nt$ are taken to form a trajectory matrix ${\bX}_{\nd} \in {\Rspace}^{3 \times \Nt} = [{\bx}_0^{\intercal},\dots,{\bx}_{\Nt}^{\intercal}]$. Trajectory matrices are reshaped into a delay-embedding vector \cite{Clainche2017}, resulting in a sample  ${\bx}_{\nd} \in {\Rspace}^{3 \Nt}$. To generate datasets, this process is repeated with random initialisations of $\q,\qdot$ and $\uv$, to form the training dataset  ${\bX}^{\Ndtrain} \in {\Rspace}^{3 \Nt \times \Ndtrain}$, validation dataset ${\bX}^{\Ndvalid} \in {\Rspace}^{3 \Nt \times \Ndvalid}$ and evaluation dataset ${\bX}^{\Ndeval} \in {\Rspace}^{3 \Nt \times \Ndeval}$. In the simulation experiment $\vf=0$, $g=-1$ and $m,l=1$.  

Using the Deep Koopman framework \sref{DKN}, a neural network architecture is learnt. For specific details on generating data, and architecture implementation details, please see Appendix.

\subsection{Rigid pendulum (no control)}
\label{s:sim_no_control}

\begin{figure*}[t!]
\includegraphics[width=0.5\linewidth]{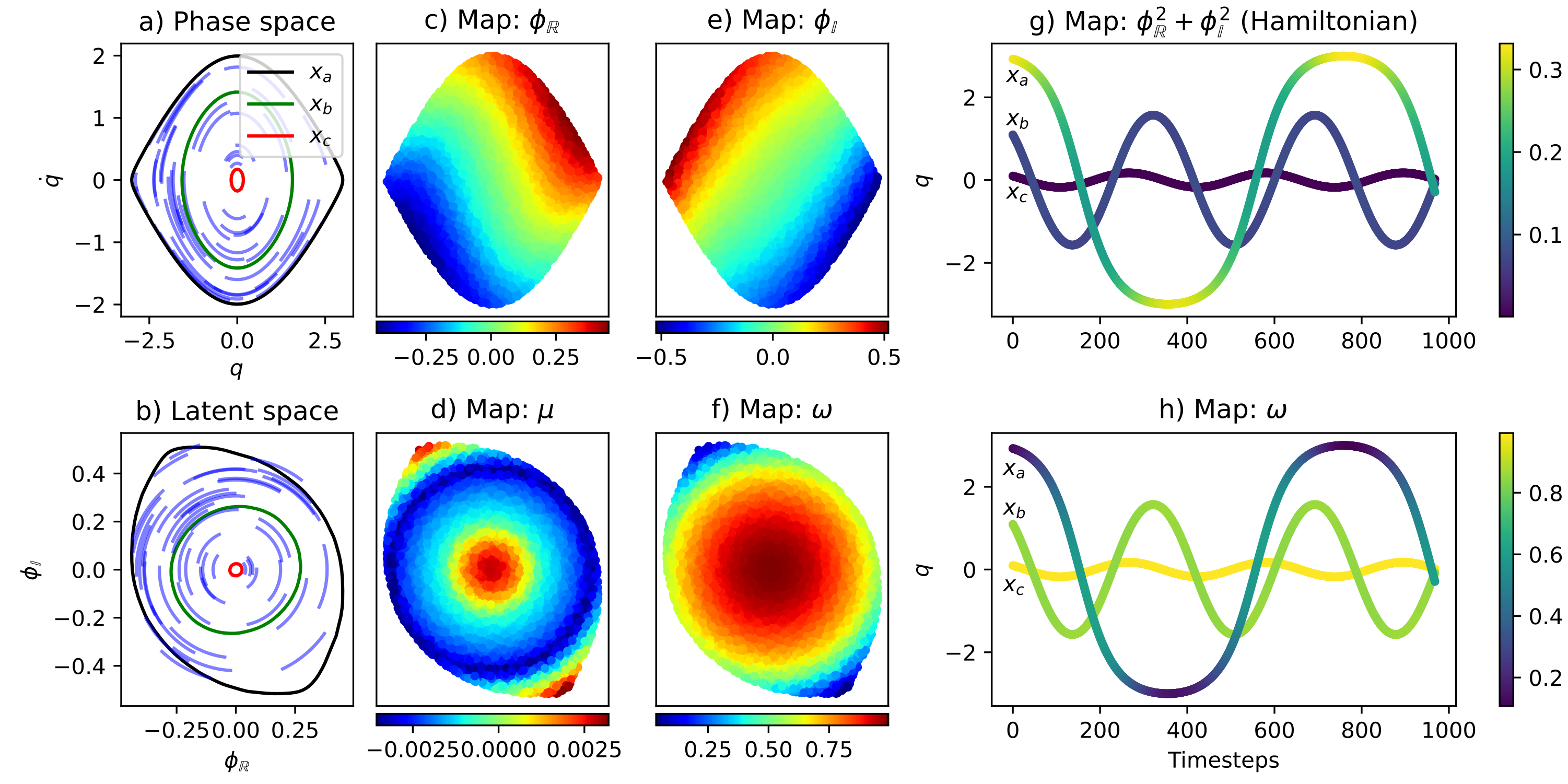}
\includegraphics[width=0.5\linewidth]{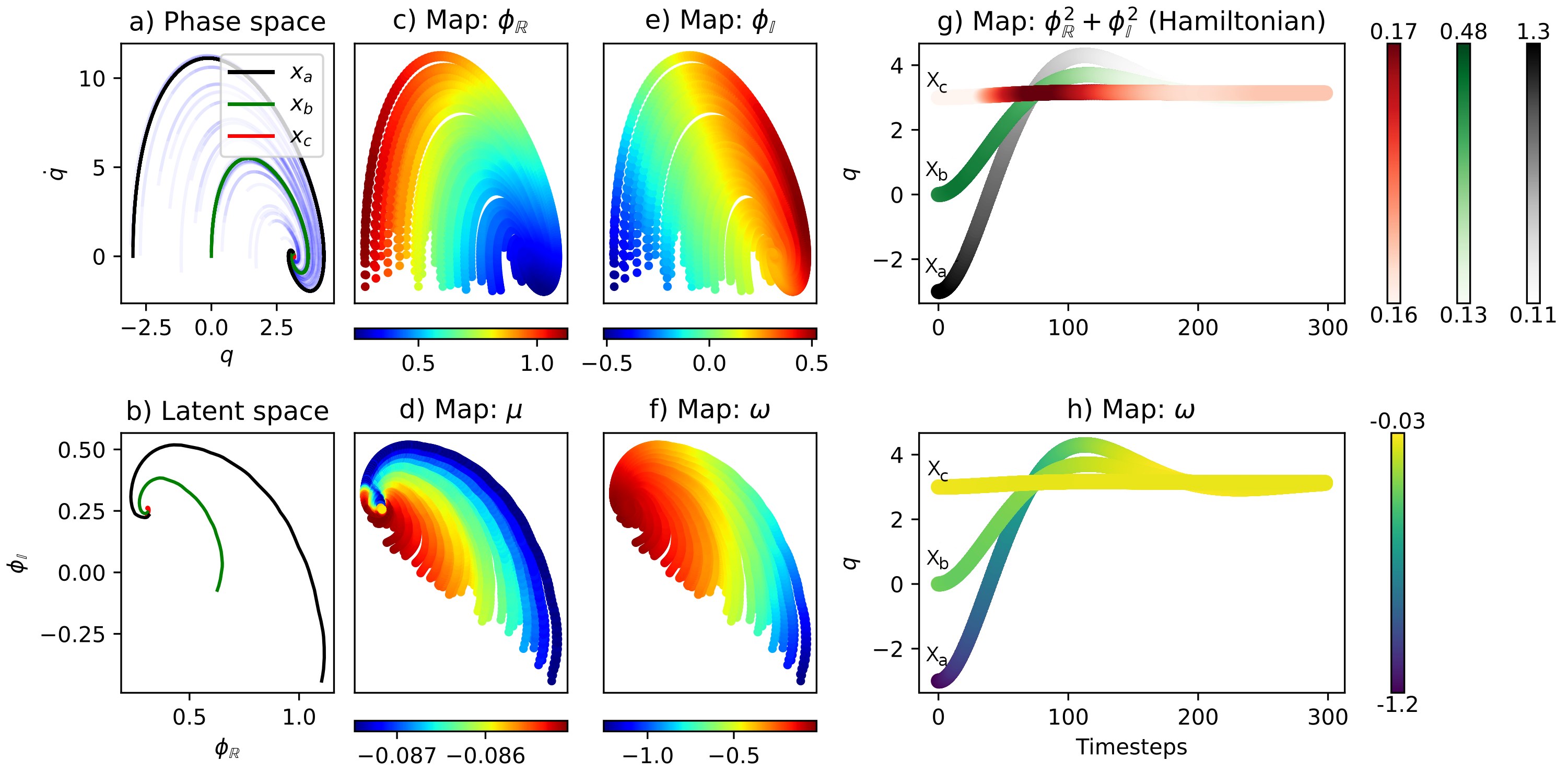}
\caption{DKN: Rigid pendulum without control (left), and with PD control (right). a) Phase space with exemplar trajectories with initial states (No control: $x_a=[\pi-0.1,0]$ , $x_b=[\frac{\pi}{2},0]$, and $x_c=[\frac{\pi}{18},0]$, PD Control: $x_a=[-\pi+0.1,0]$ , $x_b=[0,0]$, and $x_c=[\pi-0.1,0]$), b) Latent space, c,e) Single  eigenfunction with real and imaginary axes mapped independently, d) Eigenfunction growth $\KEigVReal$, f) Eigenvalue frequency $\KEigVImag$, g) Hamiltonian energy and, h) Eigenvalue frequency over time.}
\label{f:rigid_pendulum_DKN_latent}
\end{figure*}

For an initial experiment, the dynamics of a rigid pendulum without any control input is learnt. This is similar to the experiments described in \cite{lusch2018deep}, and aims to elucidate the interpretation process of the Koopman outputs.


The results for learning the dynamics with DKN are shown in \fref{rigid_pendulum_DKN_latent}  (left). In this, it is seen that as in \cite{lusch2018deep}, trajectories in the input phase space \fref{rigid_pendulum_DKN_latent}  (left) (a) are mapped via a single complex conjugate pair of learnt eigenfunctions, to a latent-space that is linear in polar coordinates \fref{rigid_pendulum_DKN_latent} (left) (b). Specifically, this demonstrates that the system is globally linearised within the Koopman framework. Additionally, by using the auxiliary network to parameterise the learnt dynamics by a continuous spectra of eigenvalues, a continuous range of oscillatory frequencies is also captured by the model,  thereby capturing the continuous eigenvalue spectrum. This is seen in the latent space (\fref{rigid_pendulum_DKN_latent} (left) (f)), where the continuous spectra captures an intuitive physical characteristic of the pendulum, that being the frequency of oscillation is dependent on the position in the phase space. Specifically, this shows the eigenvalue frequency decreases the further from the centre, \ie the period of the swing increases the further from the stable equilibrium position. In the context of the pendulum, this expressed as a high frequency oscillation at small angles, and low frequency oscillation at higher angles (\fref{rigid_pendulum_DKN_latent} (left) (h)). Additionally, characteristics such as the Hamiltonian energy \cite{lusch2018deep} can be expressed function of the phase space (\fref{rigid_pendulum_DKN_latent} (left) (g)). In the context of the pendulum this captures the kinetic and potential energy of the pendulum at top of the swing. As such, the model captures interesting inherent physical properties of the underlying dynamical system.

\subsection{Rigid pendulum (PD control)}

A subsequent experiment evaluates the performance of dynamics learning with DKN, for systems with control input. In this experiment, trajectories of motion are generated according to \sref{generating_data}, with control given by a standard PD controller:
\begin{equation}
    \uv = -K_p(\q_{target}-\q) + K_d \qdot,
\end{equation}
where $\q_{target}=\pi$ is the target, and $K_p=10$, $K_d=3$ are coefficients for the proportional and derivative terms.

In this experiment, trajectories of motion are generated for ${\Nts}=300$ timesteps, to reach the target. Trajectories are then split into segments of length ${\Nt}=50$. 


Results for this experiment are seen in \fref{rigid_pendulum_DKN_latent} (right). In this it is seen similar to \sref{sim_no_control}, a single complex eigenfunction characterises the behaviour of the dynamical system. Specifically, the frequency increases around the unstable equilibrium position \fref{rigid_pendulum_DKN_latent}  (right) (f), resulting in high frequency oscillations when the pendulum stabilises \fref{rigid_pendulum_DKN_latent} (right) (h) due to the PD gain terms. As such, it is clear that the learnt model both captures dynamics under control, and the induced PD controller dynamics.
\section{Experiment}

\subsection{Soft inverted pendulum}
Experiments in \sref{sim_experiments} demonstrate that DKN can be applied to non-linear dynamical systems under control, to obtain intuitive outputs that can be used to help explain characteristic behaviours. 

Given this, a further evaluation is performed exploring the application of this method to a more challenging scenario, stabilisation of a soft inverted pendulum in a real-environment. In the context of dynamical systems, modelling this introduces the additional challenges of both system complexity (\eg modelling non-linear elasticity and hysteresis), as well as accounting for real-world environmental noise and variations (\eg air-flow, vibrations). 

An overview of soft inverted pendulum elasticity is shown in \fref{senzai}, using an environment as shown in \fref{env}. In this, a soft inverted pendulum made from polyurethane foam (JIS ASKER C $<$ 1, $30\times100\times700$ {mm}) is rotated at the base by a robot arm (Universal Robots Inc.: UR5). The softness induces non-linear dynamics in the system, and can be represented as a type of non-linear spring with dynamics given by the Duffing equation \cite{duffing,affine}. This system has a dual-well potential, with one stable equilibrium point on either side (\eg \fref{env} demonstrates left-sided rest), and requires nonlinear control to reach a balanced unstable equilibrium point.


In these experiments, it is assumed that the dynamics can be described by the state variable $\bx = [\theta,\dot\theta,\q]^T$, where $\theta$ is the angle of pendulum, $\dot\theta$ the angular velocity of pendulum and $\q$ is the joint angle of robot. The control input $u = \qdot$ is the joint velocity of robot. Due to the hysteresis of elasticity, it is assumed that this experimental setup requires memory. As such, models use ${\Nt}=50$ time-steps as historical input, to learn the embedding and its associated linear dynamical system.

\begin{figure}[t]
\begin{center}
\includegraphics[width=0.65\hsize]{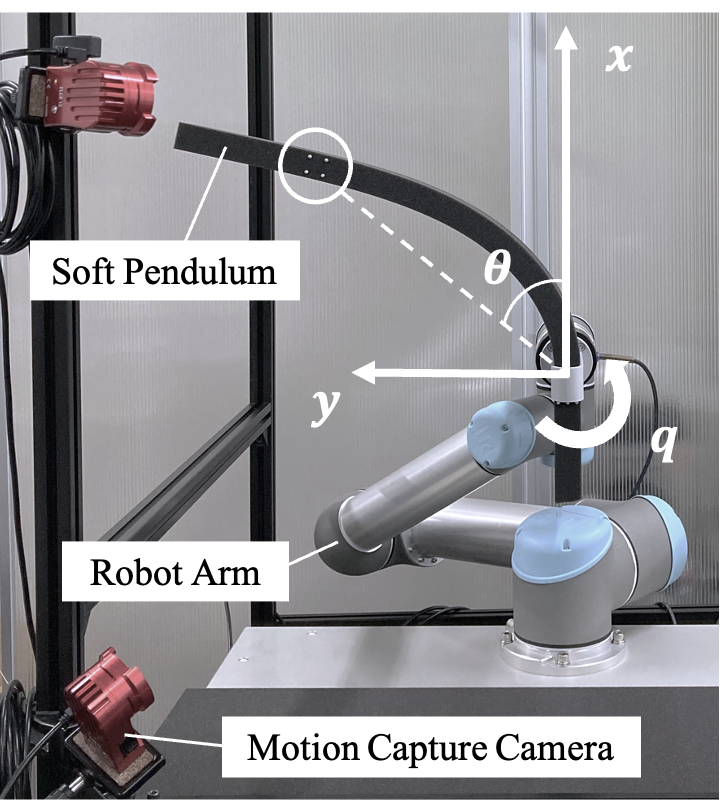}
\caption{\label{f:env} The environment of the soft inverted pendulum}
\vspace{-3mm}
\end{center}
\end{figure}

\subsection{Experimental setting}
The experimental setup is shown in \fref{env}, where the polyurethane foam prism is attached to the robot end-effector, with the center of the end-effector defined as the origin and the distance from the end of the pendulum given as $580 \text{mm}$. To measure the system state, recurrent reflection markers are attached to the end of the pendulum, and its position $p(x_t,y_t)$ is measured with a motion capture camera (OptiTrack: Flex13, sampling frequency $120$ Hz) to obtain the pendulum angle $\theta$. The angular velocity $\dot\theta$ is calculated by approximate differentiation.
\begin{align}
\theta_t &= \arctan\frac{y_t}{x_t}\\
\dot{\theta_t} &= \frac{\theta_t - \theta_{t-1}}{dt}
\label{e:mocap}
\end{align}
The state of the robot $[\q,\qdot]$ is obtained and controlled directly through the robot's internal controller.

\subsection{Data collection}
To generate trajectories of motion for dynamics learning, time series data of the soft inverted pendulum system is collected using a PD control:
\begin{equation}
    u = - K_p(\theta_{target} - \theta) + K_d\dot\theta.
    \label{e:PD}
\end{equation}
This controller acts as an exciter, to capture the dynamics of the soft pendulum, which is a non-autonomous system. Since PD control is a linear controller, the gain $K_P, K_D$ and target position $\theta_{target}$ are varied to explore the nonlinear state space, as detailed in \tref{PD}. 
Specifically, these values are chosen to generate a range of trajectories that vary between conservative trajectories that are stationary at the target point over time and unstable trajectories that oscillate by overshooting to the stable equilibrium point.

Data is collected for 30 minutes for each gain and target setting, with a control frequency of 20 Hz. 

\begin{table}[tb]
\centering
\caption{\label{t:PD} Details of PD controller for data collection}
\footnotesize
\begin{tabular}{|c|c|c|c|c|}
\hline 
Controller & PD1 & PD2 & PD3 & PD4  \\ 
\hline 
$K_p$ & 0.3 & 0.3 & 0.1 & 0.1 \\ 
\hline 
$K_d$ & 0.1 & 0.2& 0.2 & 0.3 \\ 
\hline 
$\theta_{target}$ & $0,\pm0.8$ & $0,\pm0.1$ & $0,\pm0.8$ & $0,\pm0.8$ \\ 
\hline 
\end{tabular} 
\end{table} 

\subsection{Dynamics analysis}

To learn and analyse the dynamics of the soft pendulum, a DKN network is learnt following the procedure described in \sref{DKN}, with training details given in Appendix.

\subsubsection{One Complex Eigenfunction}
\label{s:real_one_complex}

\begin{figure}[t]
\vspace{5mm}
\begin{center}
\includegraphics[width=\linewidth]{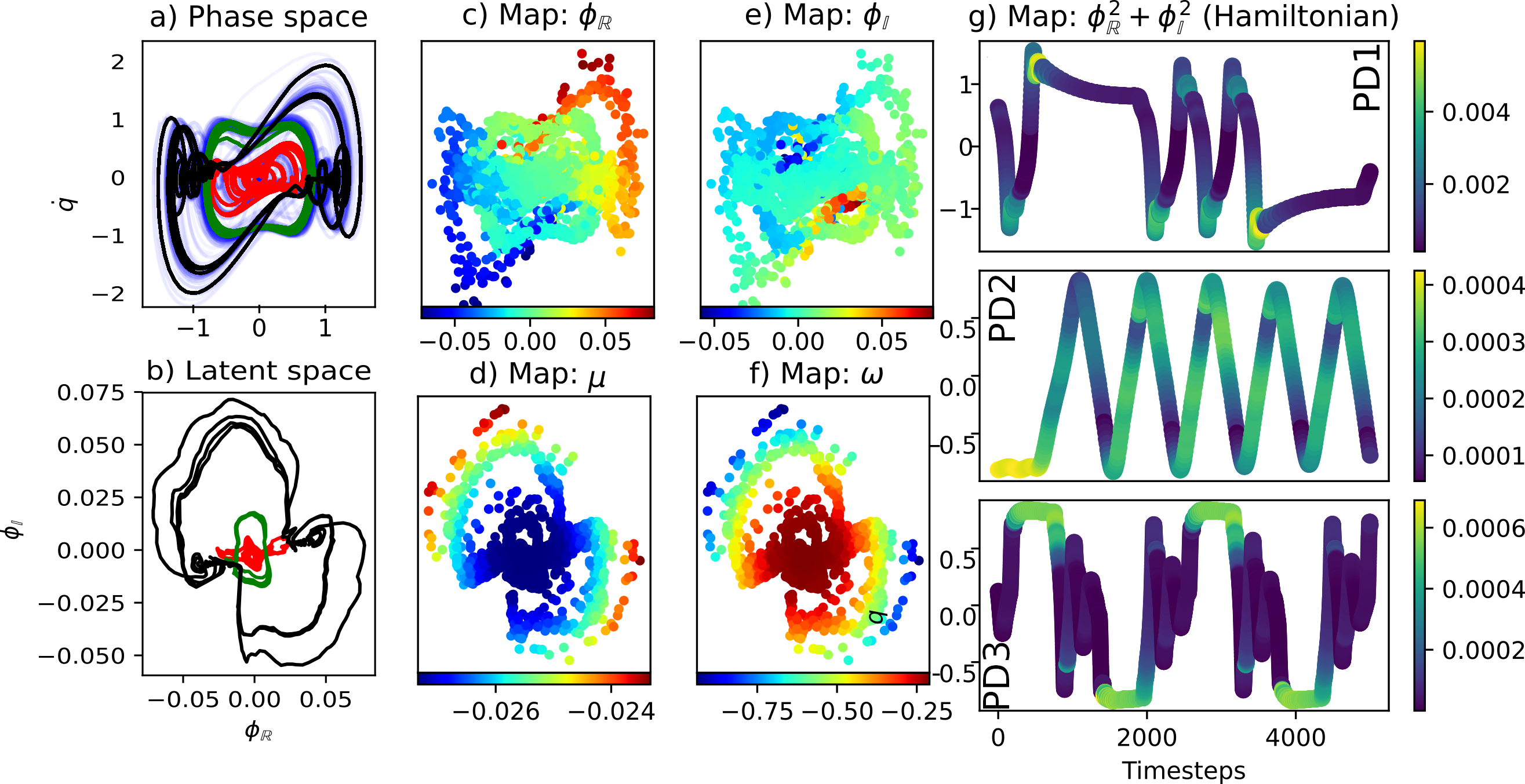}
\caption{\label{f:soft_pendulum_eig1} DKN one complex eigenfunction of the soft inverted pendulum: a) Phase space with different behaviour trajectories (for gains in \tref{PD}), b) Latent space, c,e) Single  eigenfunction with real and imaginary axes mapped independently, d) Eigenfunction growth $\KEigVReal$, f) Eigenvalue frequency $\KEigVImag$, g) Hamiltonian energy.}
\vspace{-3mm}
\end{center}
\end{figure}

As an initial experiment, the dynamics are learnt using DKN with a single complex eigenfunction (as in \sref{sim_experiments}). The results for this are seen in \fref{soft_pendulum_eig1}. In this, it is seen that the input phase space, \fref{soft_pendulum_eig1}(a), is comprised of different behavioural trajectories, due to the variations of the controller gains inducing differing oscillation periods.

As a first observation, it is seen that this variation in behaviour is similarly observed in the latent complex eigenfunction space (\fref{soft_pendulum_eig1}(b)). Specifically, input trajectories from PD1 (black), which have a longer travelling distance and contains oscillation at the dual wells, are successfull \textit{unwound} and mapped to the polar latent space. This highlights that complex behaviours in the phase space can be mapped via to the eigenfunctions to a linear space. Similarly, trajectories from the other gains are also mapped, with decreasing radious from the centre. As such, the different behaviours have been discretised into disparate regions of the eigenfunction latent space.



To explore this in the context of gaining intuative understanding of the system,  \fref{soft_pendulum_eig1}(f), shows the frequency of oscillation in the latent space. In this, the frequency varies as a function of the phase space, oscillating at high frequencies as the pendulum converges to the stabilisation point (due to the gain parameters). Additionally, as seen in \fref{soft_pendulum_eig1}(h), the energy of the system can be extracted from the model via the magnitude of the eigenfunctions. In this, it is seen that for the stabilisation trajectory (PD controller with strongest gains, red) the highest energy corresponds to the initial change in velocity.

On the other hand, $\mu$, which indicates the divergence and convergence of the oscillations in \fref{soft_pendulum_eig1}(d), takes a very small negative value, just like the rigid pendulum in \fref{rigid_pendulum_DKN_latent}.
This suggests that DKN with a single eigenfunction learns only the steady oscillations caused by the PD controller, and the stabilization dynamics of the soft pendulum up to the equilibrium point is not captured.

\subsubsection{Nine Complex Eigenfunctions}

\begin{figure}[t]
\centering
\includegraphics[width=0.9\linewidth]{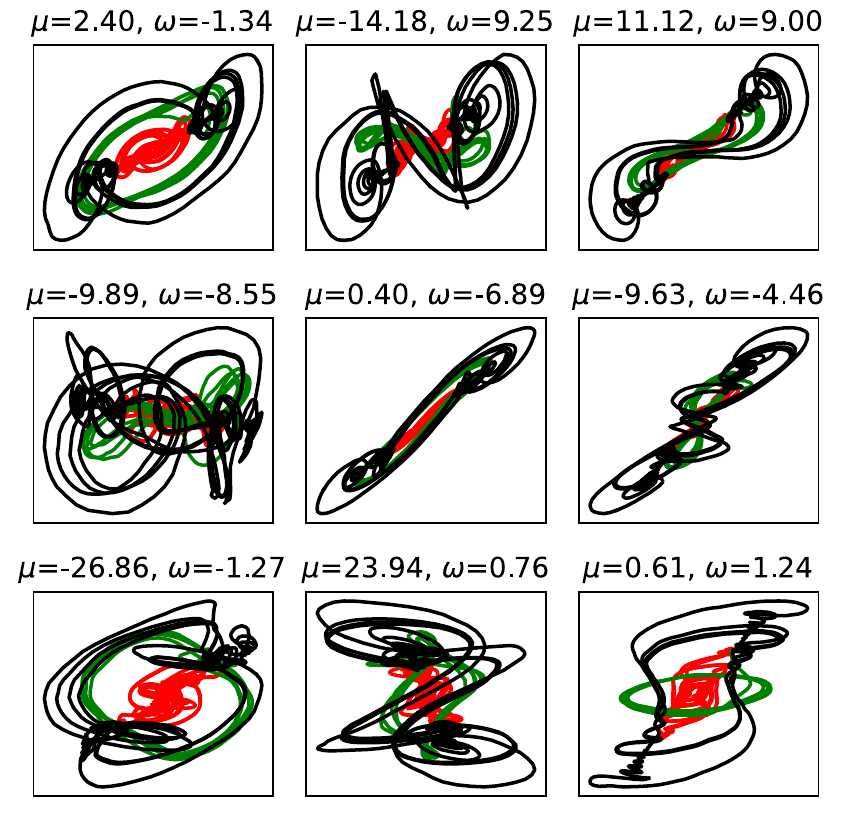}
\caption{\label{f:soft_pendulum_eig9} DKN nine complex eigenfunctions of the soft inverted pendulum: Latent coordinates for each complex eigenfunction, for exemplar trajectories given by gains in \tref{PD}}
\vspace{-3mm}
\end{figure}

In order to learn the full  dynamics, up to and including equilibrium stabilization, the experiment in \sref{real_one_complex} is repeated with a greater number of complex eigenfunctions.  Specifically, nine eigenfunctions are chosen to capture a more expansive range of dynamic components.

The results for this increased number of eigenfunctions is seen in \fref{soft_pendulum_eig9}. In this, it is seen that unique configurations are captured for each pair. However, in contrast to the experiments in \fref{soft_pendulum_eig9}, both $\mu$ and $\omega$ are learnt as discrete, non-continous values for each eigenfunction. In the context of dynamical systems analysis, this means that the learnt dynamical system can be expressed as a combination of nine fixed oscillatory behaviours, possibly corresponding to mechanical properties of the system.

\subsection{Model predictive control}
A further investigation is performed to assess the applicability of the learnt DKN models to control problems. In this, the performance is tested through MPC with CEM \eref{MPC}.
CEM is performed with the cost function:
\begin{equation}
    c({\bx},{\bx}_{target}) = w_1\theta^2 + w_2\dot\theta^2,
    \label{e:cost}
\end{equation}
where $w_1$ and $w_2$ are cost weights for pendulum angle and angular velocity, and the target state ${\bx}_{target}$ is set as $[\theta_{target}, \dot\theta_{target}] = [0,0]$. 

Trajectories of data are obtained for $30$ seconds, for $5$ samples of each PD model. The control frequency is fixed at 20 Hz, and the predictive horizon $t_p$ and the cost weights $[w_1,w_2]$ are hand-tuned.
They are evaluated for their transition and summation of the error (the cost when the weights are all one) as the achievement of control.
For video results, please see our YouTube channel \footnote{\urlstyle{rm}\url{https://youtu.be/BeXtHMoSSwM}}.

In comparison to the DKN method, a fully connected network (FCN) is simultaneously learnt using the same dataset. The FCN is formed of an encoder, decoder and latent inner hidden layers of the same network size as DKN. As such, this is a reasonable equivalent network, that lacks only the inherent linearisation of DKN.

Results for reaching the target with MPC are given in \fref{cost_step} (top), for the case of ${\Nt}=1$ (\ie no history). In this, it is seen that for DKN, the error between the tip position and target oscillates immediately from the start and is unable to stabilize for most of the time. In comparison, FCN maintains a reduced error during initial movement, but likewise destabilises and exhibits oscillatory behaviour. This is not unexpected, as due to the hysteresis of the system, historical measurements are imperative for uniquely defining the state, and failure to include this input information results in overall poor controllers. The average error and variation of each modelling approach is shown in \fref{cost_bar}. 

In comparison, by including ${\Nt}=50$ historical measurements as part of the learning and prediction process, \fref{cost_step} (bottom) shows that DKN displays a small error during runtime, immediately moving the soft pendulum tip close to the target, and maintaining stabilisation with only small error. However, including these historical measurements in FCN, does not significantly improve performance, and displays a worse stabilization performance than the ${\Nt}=1$ case. One possible reason for this, is that the complexity of modelling the relationship between such a large history, and the next step, is too high-dimensional for such a small network. However, as DKN learns both the embedding and the associated linear dynamical system, it can exploit the implicit linear bias to learn small, parsimonious models that capture the actual physics-based dynamics.

\begin{figure}[t]
\centering
\includegraphics[width=\hsize]{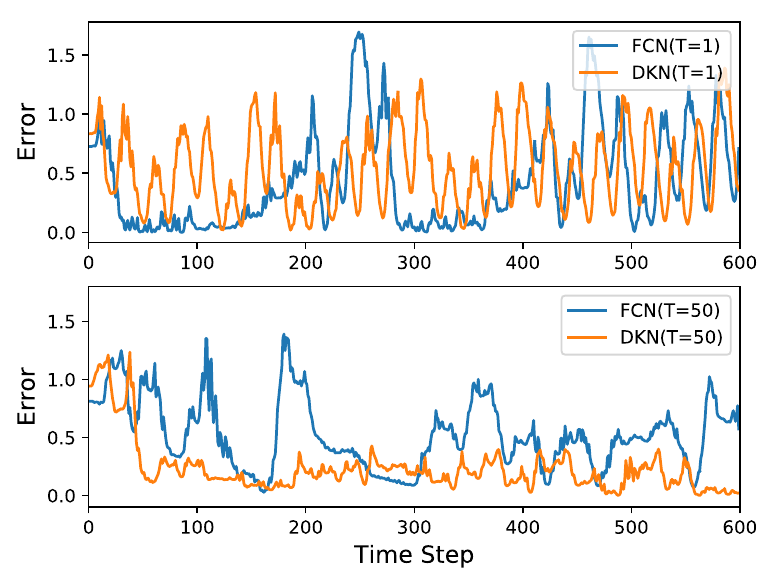}
\caption{\label{f:cost_step} Error in soft pendulum tip position during stabilization.}
\end{figure}

\begin{figure}[t]
\centering
\includegraphics[width=0.8\hsize]{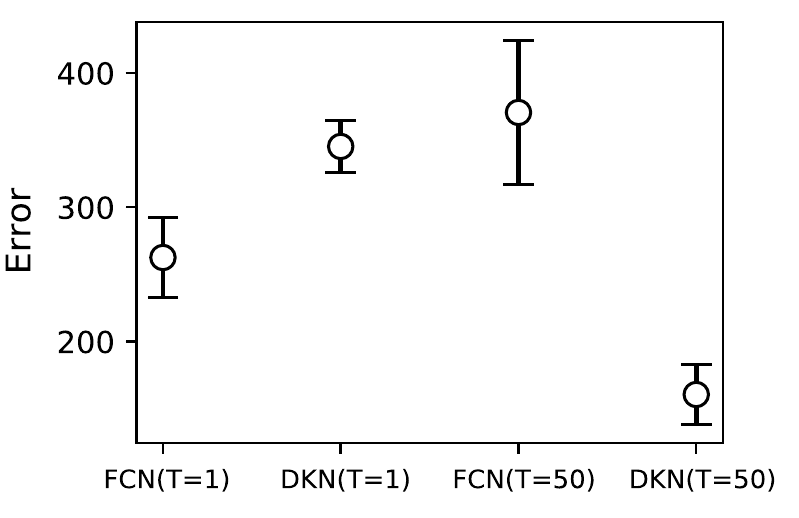}
\caption{\label{f:cost_bar} Mean sum of errors between soft pendulum position and target, with standard deviation of each method over timesteps.}
\vspace{-3mm}
\end{figure}









\section{Discussion}

This paper presents an approach to controlling and interpreting non-linear soft robotics, by learning globally linearized dynamics models. While previous studies have applied Koopman operator theory to control soft robotics, the approach in this paper explicitly analyses learnt spectral outputs to gain intuitive understanding of the systems under investigation. In combination with MPC, models are applied to control applications, resulting in models that outperform equivalent scale deep networks, due to the inherent parsimonious latent representation of dynamics. As future work, the approach will be applied to more complex soft robotics tasks, including incorporating vision for object handling.

\bibliographystyle{IEEEtran}
\bibliography{paper,paper_real_exp}

\begin{thebibliography}{10}
\providecommand{\url}[1]{#1}
\csname url@samestyle\endcsname
\providecommand{\newblock}{\relax}
\providecommand{\bibinfo}[2]{#2}
\providecommand{\BIBentrySTDinterwordspacing}{\spaceskip=0pt\relax}
\providecommand{\BIBentryALTinterwordstretchfactor}{4}
\providecommand{\BIBentryALTinterwordspacing}{\spaceskip=\fontdimen2\font plus
\BIBentryALTinterwordstretchfactor\fontdimen3\font minus
  \fontdimen4\font\relax}
\providecommand{\BIBforeignlanguage}[2]{{%
\expandafter\ifx\csname l@#1\endcsname\relax
\typeout{** WARNING: IEEEtran.bst: No hyphenation pattern has been}%
\typeout{** loaded for the language `#1'. Using the pattern for}%
\typeout{** the default language instead.}%
\else
\language=\csname l@#1\endcsname
\fi
#2}}
\providecommand{\BIBdecl}{\relax}
\BIBdecl

\bibitem{michiels2014stability}
W.~Michiels and S.~Niculescu, \emph{Stability, control, and computation for
  time-delay systems: an eigenvalue-based approach}.\hskip 1em plus 0.5em minus
  0.4em\relax SIAM, 2014.

\bibitem{ham1983observability}
F.~M. Ham and R.~G. Brown, ``Observability, eigenvalues, and kalman
  filtering,'' \emph{IEEE Transactions on Aerospace and Electronic Systems},
  no.~2, pp. 269--273, 1983.

\bibitem{grazioso2019geometrically}
S.~Grazioso, G.~Di~Gironimo, and B.~Siciliano, ``A geometrically exact model
  for soft continuum robots: The finite element deformation space
  formulation,'' \emph{Soft robotics}, vol.~6, no.~6, pp. 790--811, 2019.

\bibitem{sadati2021tmtdyn}
S.~H. Sadati, S.~E. Naghibi, A.~Shiva, B.~Michael, L.~Renson, M.~Howard, C.~D.
  Rucker, K.~Althoefer, T.~Nanayakkara, S.~Zschaler \emph{et~al.}, ``{TMTDyn}:
  A matlab package for modeling and control of hybrid rigid--continuum robots
  based on discretized lumped systems and reduced-order models,'' \emph{The
  International Journal of Robotics Research}, vol.~40, no.~1, pp. 296--347,
  2021.

\bibitem{yu1998analysis}
P.~Yu and Q.~Bi, ``Analysis of non-linear dynamics and bifurcations of a double
  pendulum,'' \emph{Journal of Sound and Vibration}, vol. 217, no.~4, pp.
  691--736, 1998.

\bibitem{kim2021review}
D.~Kim, S.~Kim, T.~Kim, B.~B. Kang, M.~Lee, W.~Park, S.~Ku, D.~Kim, J.~Kwon,
  H.~Lee \emph{et~al.}, ``Review of machine learning methods in soft
  robotics,'' \emph{PLoS One}, vol.~16, no.~2, p. e0246102, 2021.

\bibitem{heuillet2021explainability}
A.~Heuillet, F.~Couthouis, and N.~D{\'\i}az-Rodr{\'\i}guez, ``Explainability in
  deep reinforcement learning,'' \emph{Knowledge-Based Systems}, vol. 214, p.
  106685, 2021.

\bibitem{rudin2019stop}
C.~Rudin, ``Stop explaining black box machine learning models for high stakes
  decisions and use interpretable models instead,'' \emph{Nature Machine
  Intelligence}, vol.~1, no.~5, pp. 206--215, 2019.

\bibitem{huang2003neural}
J.~Huang and F.~L. Lewis, ``Neural-network predictive control for nonlinear
  dynamic systems with time-delay,'' \emph{IEEE Transactions on Neural
  Networks}, vol.~14, no.~2, pp. 377--389, 2003.

\bibitem{koopman1931hamiltonian}
B.~O. Koopman, ``Hamiltonian systems and transformation in hilbert space,''
  \emph{Proceedings of the {N}ational {A}cademy of {S}ciences of the {U}nited
  {S}tates of {A}merica}, vol.~17, no.~5, p. 315, 1931.

\bibitem{lusch2018deep}
B.~Lusch, J.~N. Kutz, and S.~L. Brunton, ``Deep learning for universal linear
  embeddings of nonlinear dynamics,'' \emph{Nature communications}, vol.~9,
  no.~1, pp. 1--10, 2018.

\bibitem{haggerty2020modeling}
D.~A. Haggerty, M.~J. Banks, P.~C. Curtis, I.~Mezi{\'c}, and E.~W. Hawkes,
  ``Modeling, reduction, and control of a helically actuated inertial soft
  robotic arm via the {K}oopman operator,'' \emph{arXiv preprint
  arXiv:2011.07939}, 2020.

\bibitem{abraham2019active}
I.~Abraham and T.~D. Murphey, ``Active learning of dynamics for data-driven
  control using {K}oopman operators,'' \emph{IEEE Transactions on Robotics},
  vol.~35, no.~5, pp. 1071--1083, 2019.

\bibitem{han2020deep}
Y.~Han, W.~Hao, and U.~Vaidya, ``Deep learning of koopman representation for
  control,'' Oct. 2020.

\bibitem{shi2022deep}
H.~Shi and M.~Q. Meng, ``Deep {K}oopman operator with control for nonlinear
  systems,'' \emph{arXiv preprint arXiv:2202.08004}, 2022.

\bibitem{mauroy2020koopman}
A.~Mauroy, I.~Mezic, and Y.~Susuki, \emph{{K}oopman Operator in Systems and
  Control}.\hskip 1em plus 0.5em minus 0.4em\relax Springer, 2020.

\bibitem{kutz2016dynamic}
J.~N. Kutz, S.~L. Brunton, B.~W. Brunton, and J.~L. Proctor, \emph{Dynamic mode
  decomposition: data-driven modeling of complex systems}.\hskip 1em plus 0.5em
  minus 0.4em\relax SIAM, 2016.

\bibitem{Korda2018}
M.~Korda and I.~Mezi{\'{c}}, ``Linear predictors for nonlinear dynamical
  systems: {K}oopman operator meets model predictive control,''
  \emph{Automatica}, vol.~93, pp. 149--160, 2018.

\bibitem{brunton2021modern}
S.~L. Brunton, M.~Budi{\v{s}}i{\'c}, E.~Kaiser, and J.~N. Kutz, ``Modern
  {K}oopman theory for dynamical systems,'' \emph{arXiv preprint
  arXiv:2102.12086}, 2021.

\bibitem{Brunton2016}
S.~L. Brunton, B.~W. Brunton, J.~L. Proctor, and J.~N. Kutz, ``{K}oopman
  invariant subspaces and finite linear representations of nonlinear dynamical
  systems for control,'' \emph{{PLOS} {ONE}}, vol.~11, no.~2, p. e0150171,
  2016.

\bibitem{kaiser2017data}
E.~Kaiser, J.~N. Kutz, and S.~L. Brunton, ``Data-driven discovery of {K}oopman
  eigenfunctions for control,'' \emph{arXiv preprint arXiv:1707.01146}, 2017.

\bibitem{bevanda2021koopman}
P.~Bevanda, S.~Sosnowski, and S.~Hirche, ``{K}oopman operator dynamical models:
  Learning, analysis and control,'' \emph{arXiv preprint arXiv:2102.02522},
  2021.

\bibitem{Shi2021}
L.~Shi and K.~Karydis, ``Acd-edmd: Analytical construction for dictionaries of
  lifting functions in {K}oopman operator-based nonlinear robotic systems,'' in
  \emph{IEEE Robotics and Automation Letters}.\hskip 1em plus 0.5em minus
  0.4em\relax IEEE, 2021.

\bibitem{mamakoukas2019local}
G.~Mamakoukas, M.~Castano, X.~Tan, and T.~Murphey, ``Local {K}oopman operators
  for data-driven control of robotic systems,'' in \emph{Robotics: science and
  systems}, 2019.

\bibitem{folkestad2020episodic}
C.~Folkestad, D.~Pastor, and J.~W. Burdick, ``Episodic {K}oopman learning of
  nonlinear robot dynamics with application to fast multirotor landing,'' in
  \emph{2020 IEEE International Conference on Robotics and Automation}.\hskip
  1em plus 0.5em minus 0.4em\relax IEEE, 2020, pp. 9216--9222.

\bibitem{bruder2020data}
D.~Bruder, X.~Fu, R.~B. Gillespie, C.~D. Remy, and R.~Vasudevan, ``Data-driven
  control of soft robots using {K}oopman operator theory,'' \emph{IEEE
  Transactions on Robotics}, vol.~37, no.~3, pp. 948--961, 2020.

\bibitem{sotiropoulos2021dynamic}
F.~E. Sotiropoulos and H.~H. Asada, ``Dynamic modeling of bucket-soil
  interactions using {K}oopman-dfl lifting linearization for model predictive
  contouring control of autonomous excavators,'' \emph{IEEE Robotics and
  Automation Letters}, vol.~7, no.~1, pp. 151--158, 2021.

\bibitem{broad2020data}
A.~Broad, I.~Abraham, T.~Murphey, and B.~Argall, ``Data-driven {K}oopman
  operators for model-based shared control of human--machine systems,''
  \emph{The International Journal of Robotics Research}, vol.~39, no.~9, pp.
  1178--1195, 2020.

\bibitem{mezic2005spectral}
I.~Mezi{\'c}, ``Spectral properties of dynamical systems, model reduction and
  decompositions,'' \emph{Nonlinear Dynamics}, vol.~41, no.~1, pp. 309--325,
  2005.

\bibitem{rowley2009spectral}
C.~W. Rowley, I.~Mezi{\'c}, S.~Bagheri, P.~Schlatter, and D.~Henningson,
  ``Spectral analysis of nonlinear flows,'' \emph{Journal of fluid mechanics},
  vol. 641, no.~1, pp. 115--127, 2009.

\bibitem{li2017extended}
Q.~Li, F.~Dietrich, E.~M. Bollt, and I.~G. Kevrekidis, ``Extended dynamic mode
  decomposition with dictionary learning: A data-driven adaptive spectral
  decomposition of the {K}oopman operator,'' \emph{Chaos: An Interdisciplinary
  Journal of Nonlinear Science}, vol.~27, no.~10, p. 103111, 2017.

\bibitem{takeishi2017learning}
N.~Takeishi, Y.~Kawahara, and T.~Yairi, ``Learning {K}oopman invariant
  subspaces for dynamic mode decomposition,'' in \emph{Advances in Neural
  Information Processing Systems}, 2017, pp. 1130--1140.

\bibitem{kamb2020time}
M.~Kamb, E.~Kaiser, S.~L. Brunton, and J.~N. Kutz, ``Time-delay observables for
  {K}oopman: Theory and applications,'' \emph{SIAM Journal on Applied Dynamical
  Systems}, vol.~19, no.~2, pp. 886--917, 2020.

\bibitem{peter2013generalized}
T.~Peter and G.~Plonka, ``A generalized prony method for reconstruction of
  sparse sums of eigenfunctions of linear operators,'' \emph{Inverse Problems},
  vol.~29, no.~2, p. 025001, 2013.

\bibitem{cem}
P.~de~Boer, D.~P. Kroese, S.~Mannor, and R.~Y. Rubinstein, ``A tutorial on the
  cross-entropy method,'' \emph{Annals of Operations Research}, vol. 134,
  no.~1, pp. 19--67, Feb. 2005.

\bibitem{doya2000reinforcement}
K.~Doya, ``Reinforcement learning in continuous time and space,'' \emph{Neural
  computation}, vol.~12, no.~1, pp. 219--245, 2000.

\bibitem{Clainche2017}
S.~Le~Clainche and J.~M. Vega, ``Higher order dynamic mode decomposition,''
  \emph{SIAM Journal on Applied Dynamical Systems}, vol.~16, no.~2, pp.
  882--925, 2017.

\bibitem{duffing}
G.~Litak, M.~Coccolo, M.~I. Friswell, S.~F. Ali, S.~Adhikari, A.~W. Lees, and
  O.~Bilgen, ``Nonlinear oscillations of an elastic inverted pendulum,'' in
  \emph{2012 IEEE 4th International Conference on Nonlinear Science and
  Complexity}, Aug. 2012, pp. 113--116.

\bibitem{affine}
C.~D. Santina, ``The soft inverted pendulum with affine curvature,'' in
  \emph{2020 59th IEEE Conference on Decision and Control}, Dec. 2020, pp.
  4135--4142.

\end{thebibliography}

\section*{Appendix}
\subsection*{Data collection}

\hspace{-4mm}
\begin{tabular}{|c|c|c|c|}
\hline 
• & \thead{Pendulum\\ (no control)} & \thead{Pendulum\\ (PD control)} & \thead{Soft pendulum} \\ 
\hline 
$\Nt$ & 50 & 50 & 50 \\ 
\hline 
dt & 0.02 & 0.01 & 0.05 \\ 
\hline 
$\q_0$ range & [-3.1,3.1] & [-3.1,3.1] & [-1.5,1.5] \\ 
\hline 
$\qdot_0$ range & [-2,2] & [-2,2] & [-2,2] \\ 
\hline 
\# Training & 15000 & 15950 & 100763 \\ 
\hline 
\# Validation & 1000 & 1450 & 28791 \\ 
\hline 
\# Evaluation & 3000 & 2900 & 14394 \\ 
\hline 
\end{tabular}

\end{document}